\documentclass[10pt]{article}
\usepackage[utf8]{inputenc}
\usepackage[T1]{fontenc}
\usepackage{arxiv}
\usepackage{amsmath,amssymb}
\usepackage{graphicx}
\usepackage{array,booktabs,multirow,tabularx}
\usepackage{enumitem}
\usepackage{microtype}
\usepackage{marvosym}
\usepackage{fontawesome5}
\usepackage{tcolorbox}
\usepackage[numbers,sort&compress]{natbib}
\usepackage{xcolor}
\usepackage{url}
\usepackage{hyperref}
\usepackage[font=small,labelfont=bf]{caption}

\definecolor{linkblue}{RGB}{36,75,110}
\hypersetup{colorlinks=true,linkcolor=linkblue,citecolor=linkblue,urlcolor=linkblue,
  pdftitle={OpenZilo: An Open Smart-Ring Platform for Human-Centric Sensing and Interaction with Robots},
  pdfauthor={Qianggang Ding; Xingyao Wang; Rui Feng; Chi Qin; Weijian Lin; Lizhi Ding; Jiankai Tang; Yuntao Wang; Xiangke Liu; Bang Liu}}
\renewcommand{\shorttitle}{%
  {\footnotesize\itshape OpenZilo: An Open Smart-Ring Platform for Human-Centric Sensing and Interaction with Robots}}
\renewcommand{\headeright}{}
\renewcommand{\undertitle}{Preprint}
\setlist{itemsep=2pt,topsep=4pt}
\date{}

\title{OpenZilo: An Open Smart-Ring Platform for Human-Centric Sensing and Interaction with Robots}

\begin{document}
% Zilo Lab title card. Original logo: figures/zilo-logo.svg.
\definecolor{cardbackground}{HTML}{F3F3F3}
\definecolor{cardink}{HTML}{171717}
\definecolor{cardmuted}{HTML}{626262}
\definecolor{cardaccent}{HTML}{444444}
\definecolor{cardbrandlab}{HTML}{707070}
\thispagestyle{empty}
\begingroup
\colorlet{linkblue}{cardink}
\hypersetup{linkcolor=cardink,citecolor=cardink,urlcolor=cardink}
\setlength{\parskip}{0pt}
\begin{tcolorbox}[colback=cardbackground,colframe=cardbackground,coltext=cardink,
  boxrule=0pt,arc=12pt,outer arc=12pt,boxsep=0pt,
  left=22pt,right=22pt,top=22pt,bottom=22pt,
  width=\textwidth,before skip=0pt,after skip=0pt]
\begin{minipage}[t][\dimexpr0.98\textheight-44pt\relax][t]{\linewidth}
{\sffamily\fontsize{12}{15}\selectfont\bfseries Zilo\,{\color{cardbrandlab}Lab}\par}
\vspace{19pt}
{\sffamily\bfseries\fontsize{17}{21}\selectfont\raggedright
OpenZilo: An Open Smart-Ring Platform for\\
Human-Centric Sensing and Interaction with Robots\par}
\vspace{13pt}
{\sffamily\fontsize{10}{14}\selectfont\raggedright
\href{https://orcid.org/0000-0001-8415-0476}{\textbf{Qianggang Ding}}\textsuperscript{1,2,8,\Letter},
\href{https://orcid.org/0000-0002-9502-4472}{\textbf{Xingyao Wang}}\textsuperscript{3,8},
\href{https://orcid.org/0000-0002-2422-1786}{\textbf{Rui Feng}}\textsuperscript{4,8},
\href{https://orcid.org/0009-0005-0106-3497}{\textbf{Chi Qin}}\textsuperscript{8},
{\color{linkblue}\textbf{Weijian Lin}}\textsuperscript{8},\\
\href{https://orcid.org/0009-0009-3557-9837}{\textbf{Lizhi Ding}}\textsuperscript{5,8},
\href{https://orcid.org/0009-0009-5388-4552}{\textbf{Jiankai Tang}}\textsuperscript{6},
\href{https://orcid.org/0000-0002-4249-8893}{\textbf{Yuntao Wang}}\textsuperscript{6},
{\color{linkblue}\textbf{Xiangke Liu}}\textsuperscript{7},
\href{https://orcid.org/0000-0002-9483-8984}{\textbf{Bang Liu}}\textsuperscript{1,2,\Letter}\par}
\vspace{7pt}
{\sffamily\fontsize{8}{11}\selectfont\color{cardmuted}\raggedright
\textsuperscript{1}Universit\'e de Montr\'eal\\
\textsuperscript{2}Mila -- Quebec Artificial Intelligence Institute\\
\textsuperscript{3}Southeast University\quad
\textsuperscript{4}Nanjing Medical University\\
\textsuperscript{5}Dalian University of Technology\quad
\textsuperscript{6}Tsinghua University\\
\textsuperscript{7}DeepMirror\quad\textsuperscript{8}Zilo Lab\\[3pt]
\textsuperscript{\Letter}Corresponding authors\par}
\vspace{20pt}
{\rmfamily\fontsize{10.5}{13.5}\selectfont
Wearable robot interfaces must connect user input to timely physical action. We present the proposed platform, an open smart-ring platform for studying this connection across task stages. The ring integrates a microphone, an inertial measurement unit, photoplethysmography, and skin-temperature sensing. Its software architecture separates acquisition, event interpretation, interaction policy, and robot execution. We characterize seven operating modes and evaluate speech and gesture control with 10 participants. Across five commands and 250 trials per modality, speech and gestures achieved task success rates of 90.4\% and 88.8\%, respectively. Recognition accuracy exceeded task success for both modalities. Gaussian hidden Markov models achieved higher accuracy than a support vector machine across three vocabulary sizes, with longer scoring times. Following a shared spoken task start, gesture-based adjustments had shorter median response times than speech. Whole-task success was 70.0\% with gestures and 64.0\% with speech. We further examine physiological-event-triggered robot interaction using simulated event notifications. Mean notification-receipt-to-arrival times were 11.71 s for the robot and 9.20 s for an on-site person. These descriptive findings support evaluating wearable robot interfaces through both recognition and physical outcomes, with timing measures appropriate to each task stage.\par}
\vspace{13pt}
{\sffamily\fontsize{8}{11}\selectfont\color{cardmuted}
\par}
\vfill
\vspace{15pt}
\noindent\begin{minipage}[b]{0.72\linewidth}
  \sffamily\fontsize{8.5}{12}\selectfont
  \newcommand{\cardicon}[1]{\makebox[12pt][l]{\color{cardaccent}#1}}
  \cardicon{\faIcon{calendar-alt}}\textbf{Date:} November 1, 2026\\
  \cardicon{\faIcon{envelope}}\textbf{Correspondence:}
  \href{mailto:qianggang.ding@umontreal.ca}{qianggang.ding@umontreal.ca},\\
  \hspace*{12pt}\href{mailto:bang.liu@umontreal.ca}{bang.liu@umontreal.ca}\\
  \cardicon{\faIcon{github}}\textbf{Code:}
  \href{https://github.com/ziloai/openzilo}{github.com/ziloai/openzilo}\\
  \cardicon{\faIcon{globe}}\textbf{Website:}
  \href{https://openzilo.com}{openzilo.com}
\end{minipage}\hfill
\begin{minipage}[b]{0.24\linewidth}
  \raggedleft
  \includegraphics[width=42pt]{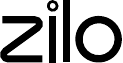}
\end{minipage}
\end{minipage}
\end{tcolorbox}
\endgroup

\clearpage

\section{Introduction}
\label{sec:introduction}
Robot interaction requires translating a user's intention into actions supported by the robot and its environment. Affordance-grounded skill selection connects language instructions to executable actions~\cite{brohan2023can}. Once execution begins, feedback and further human input can change the appropriate response~\cite{huang2022inner}. In shared spaces, users may need to redirect a task, stop motion, and verify what the robot actually did. An interaction interface must therefore support intervention and physical feedback alongside task specification.

Input requirements vary across task stages. Speech can specify a task, whereas a hand motion can provide an adjustment during execution. Recognizing an isolated command does not establish that the intended physical action occurred or that a later turn or stop was timely. Photoplethysmography (PPG) and skin-temperature sensing introduce a further possibility: contextual events that request bounded task adaptation. Such events require independent validation, while their physical effects require separate evaluation.

\begin{figure}[t] 
  \centering
  \includegraphics[width=\textwidth]{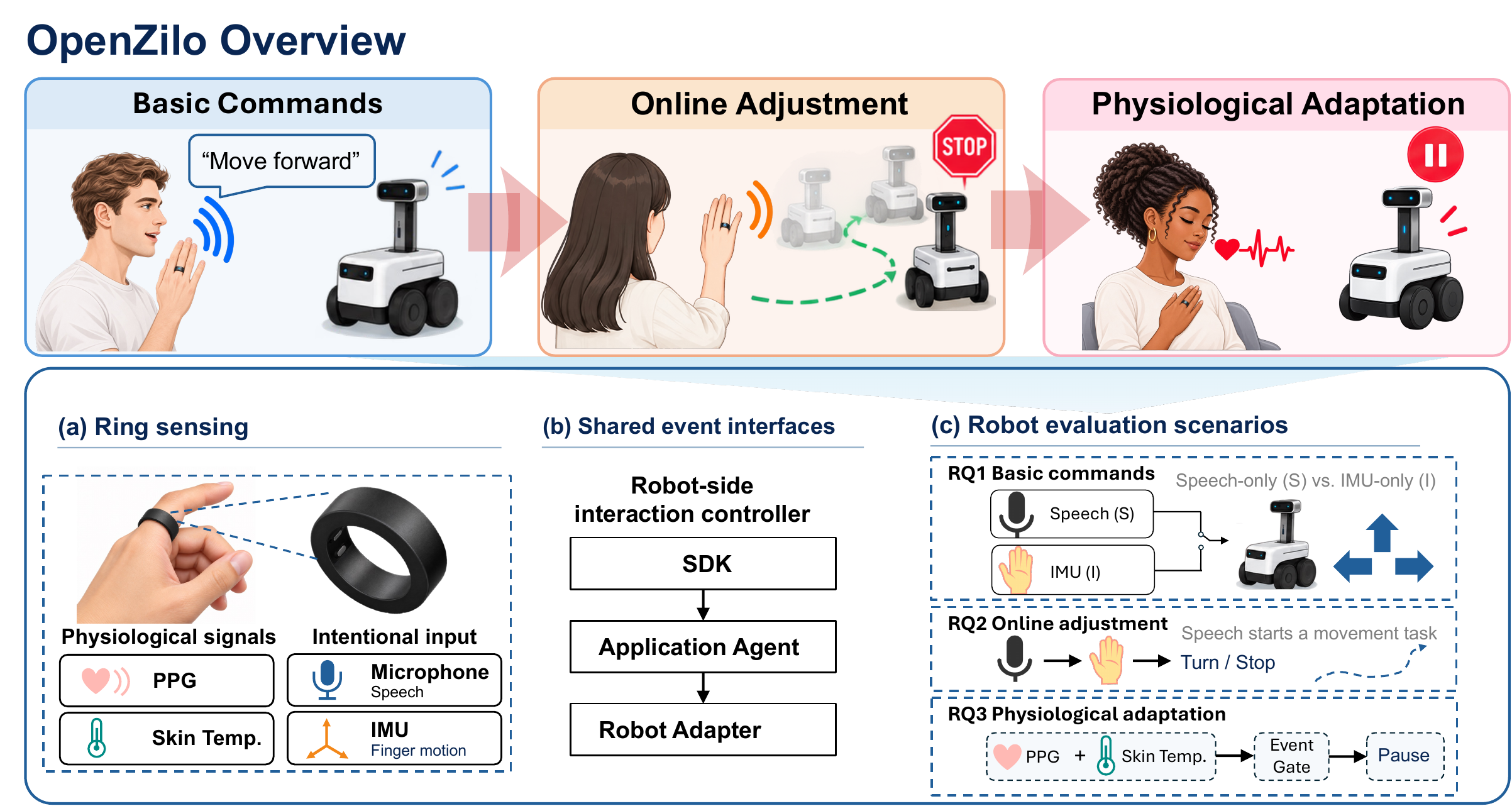}
    \caption{\textbf{Overview of the proposed platform and its robot-interaction scenarios.} The upper panels illustrate basic commands, online adjustments, and physiological-event-triggered interaction. The lower panels show (a) the four sensing modalities, (b) the SDK, application agent, and robot adapter interfaces, and (c) interaction scenarios corresponding to RQ1--RQ3. The physiological-event branch illustrates the interaction concept; RQ3 evaluates local-arrival responses to simulated physiological-event notifications.}
  \label{fig:introduction}
\end{figure}

An open interaction platform makes the path from wearable input to robot action inspectable. Hardware designs and firmware support reproduction, while timestamped streams, configurations, and software development kit (SDK) events support analysis of input interpretation and execution. Open rings already provide configurable physiological and inertial acquisition~\cite{tang2025ring}. For robot interaction, acquisition and recognition measures alone do not establish whether an input produces the intended physical response. The proposed platform addresses this evaluation gap through interfaces connecting acquisition, application events, robot commands, and physical feedback. Figure~\ref{fig:introduction} summarizes the architecture and interaction scenarios.

The proposed platform comprises the smart ring hardware and a software architecture separating acquisition, event interpretation, interaction policy, and robot execution. The microphone and inertial measurement unit (IMU) support explicit speech and gesture commands. PPG and skin temperature provide sensing channels for future physiological-context studies. This separation allows researchers to examine how recognized inputs translate into physical outcomes.

We evaluate bounded robot routines at three interaction stages. Isolated commands test recognition and execution; a continuous movement task tests turning and stopping after a shared spoken start. Physiological-event-triggered interaction examines how event notifications initiate robot responses, using simulated scenarios to compare robot and on-site human response times.

We ask three questions:
\begin{itemize}
  \item \textbf{RQ1: Single-modality commands.} How do speech and IMU gestures compare in recognition accuracy, task success, and response time for bounded robot actions? How does gesture vocabulary size affect offline recognition performance and model-scoring time?
  \item \textbf{RQ2: Online adjustment.} After the same spoken task start, how do speech and gestures compare for online turning and stopping in intervention success, response time, and whole-task success?
  \item \textbf{RQ3: Physiological-event-triggered robot interaction.} How can physiological-event notifications trigger robot actions, and how do the response times of a robot and an on-site person compare under simulated physiological-event scenarios?
\end{itemize}

The contributions of this work are threefold:
\begin{enumerate}[label=\arabic*)]
    \item We present a four-sensor ring architecture and provide electrical characterization across seven operating modes.
    \item We design a modular system architecture that separates sensor acquisition, event interpretation, task policy, and robot feedback.
    \item We evaluate command execution, gesture-vocabulary scaling, online adjustment, and physiological-event-triggered robot responses under simulated scenarios. These evaluations connect command and event inputs to physical task outcomes.
\end{enumerate}

\section{Related Work}
\label{sec:related-work}
\subsection{Rings as Interaction and Research Platforms}
Smart-ring research connects finger-worn sensing to both interactive applications and reusable research infrastructure. IRIS combines a camera, IMU, and button for device identification and context-dependent smart-home control~\cite{kim2024iris}. Users point and click to select a device, while hand rotation supports adjustments such as television volume. Smartphone-based image processing uses object recognition and scene context to distinguish device instances. This coupling of target identification and input interpretation supports actions adapted to the surrounding environment.

Other ring designs expand the range of observable hand movements and contact interactions. Ring-a-Pose uses acoustic sensing in a single untethered ring for continuous 3D hand-pose tracking and micro-gesture recognition~\cite{yu2024ring}. DualRing combines thumb- and index-worn IMUs with contact detection to capture hand motion and relative finger movement~\cite{liang2021dualring}. These signals support within-hand, hand-to-surface, and hand-to-object interactions. Addressing the power and size constraints of finger-worn input, picoRing pairs a passive ring with a wristband reader for battery-free thumb-to-index interaction~\cite{takahashi2024picoring}. Together, these systems illustrate how sensing modality, placement, and supporting hardware shape a ring's input capabilities.

Research platforms expose sensing resources for applications beyond a fixed interaction technique. OmniRing combines IMU and PPG sensing in an extensible wireless ring, demonstrating finger-motion tracking and heart-rate estimation~\cite{zhou2023omniring}. $\tau$-Ring provides synchronized PPG, six-axis IMU, and temperature acquisition, with configurable sampling rates, power modes, and wireless protocols~\cite{tang2025ring}. Its open-source Android suite supports live streaming and offline recording, with demonstrations in heart-rate monitoring and handwriting recognition. These platforms establish multisensor acquisition, configurable operation, and accessible software as foundations for reusing ring hardware across behavioral and physiological studies. At the interpretation layer, HMotionGPT aligns smart-ring inertial signals with natural language for hand-centric activity understanding~\cite{gao2026hmotiongpt}. It supports activity classification and motion descriptions, extending the use of ring signals beyond predefined gesture labels.

The proposed platform focuses on how ring-derived events translate into physical robot actions. The microphone and IMU support explicit commands, while PPG and skin temperature remain available for future physiological-context studies. The SDK, application agent, and robot adapter separate data access, event interpretation, action selection, and execution feedback. This organization supports studying how recognition and interaction policies jointly affect physical responses. Sections~\ref{sec:platform}--\ref{sec:evaluation} connect hardware characterization with this application-level evaluation.

\subsection{Speech and Gesture for Robot Interaction}
Finger-worn sensing supports both discrete gestures and continuous input. FingerSound combines a contact microphone and gyroscope in a thumb-mounted ring to recognize unistroke thumb gestures for eyes-free input~\cite{zhang2017fingersound}. ActualTouch detects contact through finger microvibrations and surface-friction vibrations, providing a touch state that can delimit gesture strokes~\cite{shi2020ready}. MouseRing combines recurrent modeling with motion-aware physical constraints for touchpad-like pointing~\cite{shen2024mousering}. WritingRing reconstructs handwriting trajectories from a single ring IMU~\cite{he2025writingring}. These techniques use motion characteristics and interaction constraints to recover meaningful input.

Reliability during continued use introduces requirements beyond isolated gesture classification. VibRing recognizes subtle single-handed gestures through a finger-worn vibroacoustic sensor~\cite{li2025vibring}. Its evaluations span multiple days and periodic gesture input during office tasks, examining recognition beyond a single controlled session. Free-air command interfaces also need to distinguish intentional gestures from intervening movements. Lee and Kim address gesture spotting with a threshold model based on hidden Markov models (HMMs), rejecting non-gesture patterns in continuous motion~\cite{lee1998spotting}. For robot control, gesture boundaries and unintended activations therefore matter alongside classification accuracy.

Language and gesture contribute complementary information to robot instructions. Tellex et al.'s Generalized Grounding Graphs relate natural-language commands to objects, places, and actions for navigation and manipulation~\cite{tellex2011understanding}. Matuszek et al. combine language with deictic gestures to identify the object a user refers to~\cite{matuszek2014learning}. Pointing helps resolve a referent that language alone may leave ambiguous. These studies establish semantic grounding as one role for multimodal input, with different channels jointly specifying the intended target or action.

Gesture interfaces also support direct robot motion control. Yau et al. examined combinations of force, touch, and inertial input for one-handed drone control, considering performance, task load, and interface size~\cite{yau2020subtle}. HyperPalm combines five static hand gestures with six-degree-of-freedom hand motion to control a quadruped's position and orientation~\cite{nazarova2022hyperpalm}. TeLeMan integrates wearable IMU-based motion capture with visual telepresence for whole-body control of a legged manipulator~\cite{zhou2022teleman}. These systems connect human movement to robot motion, making control mappings and execution feedback central to interaction.

Evaluating input methods also requires attention to task outcomes and user experience. Campagna et al. compared combined gesture-and-voice trajectory control with a smartPAD in industrial human--robot collaboration~\cite{campagna2025fostering}. Their evaluation considered task execution time, cognitive workload, and trust. Such measures complement recognition accuracy by characterizing the consequences of input choice during a complete task.

Intervention during execution introduces a further control requirement. Sheridan's account of supervisory control emphasizes task specification, monitoring, and intervention while automation carries out a task~\cite{sheridan2016human}. LILAC uses language to condition a learned low-dimensional control space for shared autonomy~\cite{cui2023lilac}. Users guide manipulation within that space, and subsequent language corrections refine it during execution. Language thus supports ongoing correction as well as initial instruction. Evaluating such interaction requires attention to how an input changes the current task and when that change takes physical effect.

The proposed platform focuses on speech and ring gestures within predefined robot routines. RQ1 assigns the modalities to identical bounded actions and separately examines gesture-vocabulary scaling. RQ2 fixes the spoken task start and compares modalities for subsequent turning and stopping. The evaluation distinguishes recognition from physical execution and separates post-input latency from cue-to-effect response time. This design examines how each input method supports changes to robot behavior at different task stages.

\subsection{User-State-Aware and Event-Triggered Robot Interaction}
Robot behavior can adapt to user-state cues as well as explicit commands. Bagheri et al.'s cognitive empathy model selects empathic responses from facial expressions and user personality~\cite{bagheri2020autonomous}. The Ryan robot combines facial-expression recognition with speech sentiment analysis to select affective dialogue and mirror positive expressions~\cite{abdollahi2023artificial}. VITA adapts across repeated mental well-being coaching sessions, using facial valence and speech duration for personalization and dialogue context for conversational actions~\cite{spitale2025vita}. These systems connect observed user behavior to communicative responses, motivating consideration of how inferred state can also inform physical task execution.

Physiological sensing introduces both opportunities for adaptation and requirements for validating the inferred state. FuSenseRing combines PPG, electrocardiography (ECG), skin temperature, and contact-force sensing for cuffless blood-pressure estimation~\cite{li2025fusensering}. Its temperature-adaptive fusion addresses thermal effects on physiological signals, illustrating why sensing conditions matter for inference. Connecting physiological inference to robot behavior requires additional evaluation. Rani and Sarkar infer operator engagement from physiological signals and demonstrate robot adaptation in an open-loop experiment that replays recorded signals at real-time speed~\cite{rani2007engagement}. cStress evaluates stress inference from ECG and respiration against laboratory protocol labels and field self-reports, with activity screening~\cite{hovsepian2015cstress}. Its sensor configuration differs from ring-based PPG and temperature acquisition, so its evidence informs validation methodology without establishing ring-based inference performance. Together, these studies motivate separating state estimation, response-policy behavior, and interaction outcomes.

Discrete health-related events introduce a complementary response path. Hobbit integrates fall and emergency detection with behavior coordination and calls to relatives or emergency services~\cite{bajones2018hobbit}. Its deployment also highlights the need to manage interruptions across ongoing activities. Bagal\`a et al. found that accelerometer-based algorithms developed using simulated falls performed less well on real-world falls~\cite{bagala2012fall}. Complementing this algorithm-level evidence, Casilari and Silva compared real-world and simulated fall datasets and found differences in their acceleration patterns~\cite{casilari2022analytical}. These findings limit what simulated-fall evaluations can establish about real-world detection. For event-triggered robot responses, detection validity, alert delivery, and physical arrival therefore require separate assessment, each with its own failure modes and timing.

Ding et al. extend embodied intelligence to personal contexts through ComBodied Agents~\cite{ding2026combodied}. Their framework organizes multimodal perception and intervention around a person's evolving state and agency, with wearables and robots providing ongoing support. The proposed platform addresses the interface between sensing, explicit user input, and bounded robot responses within this broader design space. RQ3 examines physiological-event-triggered robot interaction at the interface between event notifications and physical responses. Simulated physiological-event notifications initiate a local-arrival task, allowing comparison of robot and on-site human response times.

\section{The Proposed Platform and Open Hardware}
\label{sec:platform}

\subsection{Design Goals and Platform Architecture}

OpenZilo is an open wearable platform designed to connect multimodal finger-worn sensing with robot interaction. The platform separates four components: ring sensing, a host SDK, an application agent, and a robot SDK adapter. The smart ring provides the sensing hardware; the host SDK exposes sensor data and device capabilities. The application agent interprets recognized events according to task context, while the robot adapter translates action requests into robot-specific commands and returns execution feedback. This modular design allows interaction policies, recognition methods, and robot platforms to evolve independently of the underlying ring hardware and communication interface.

Three design goals guide the OpenZilo architecture.

\noindent\textbf{Support interaction across task stages.}
OpenZilo is designed to support task initiation, online adjustment, and event-triggered interaction. Speech enables explicit task specification, while gestures provide an alternative input channel for direct commands and adjustments during execution. Physiological signals introduce the possibility of context-aware interaction, in which validated physiological events may trigger robot responses. The application agent determines which events are actionable according to the current task state. The evaluations in RQ1--RQ3 examine different interaction stages without requiring every sensing modality to operate simultaneously.

\noindent\textbf{Separate sensing, interaction policy, and robot execution.}
Sensor acquisition and recognition are decoupled from application-level decisions and robot-specific execution. Recognizers produce application events, the agent applies task-state rules, and the robot adapter converts authorized requests into executable commands. This separation enables researchers to replace recognition algorithms, interaction policies, or robot platforms while preserving a common sensing interface. Changes to individual components still require verification of event semantics and physical execution outcomes.

\noindent\textbf{Make input-to-action behavior traceable.}
The platform links timestamped sensor records, recognized events, agent decisions, command requests, and observed robot states. Such traces help distinguish recognition errors, policy rejection, communication failures, and unsuccessful physical execution. Response time is measured against the physical outcome relevant to each task, such as motion onset for movement commands or verified stationary state for stopping.

\subsection{Integrated Multisensor Hardware}
\label{sec:ring-hardware}

\begin{figure}[t]
    \centering
    \includegraphics[width=1.0\textwidth]{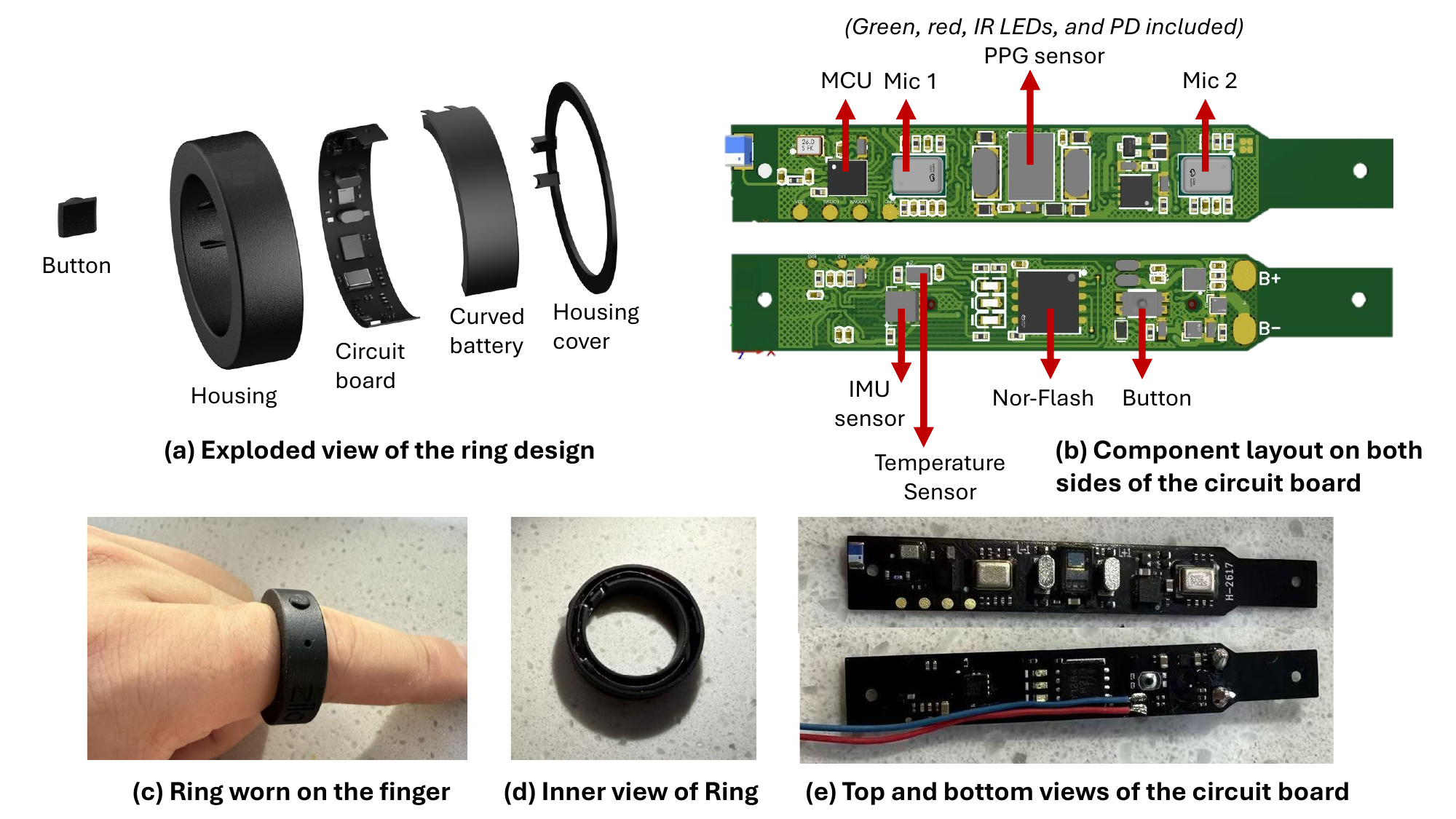}
    \caption{\textbf{Design and implementation of OpenZilo.}
    (a) Exploded view showing the button, ring housing, circuit board, curved battery, and housing cover.
    (b) Component layout on both sides of the circuit board, including the MCU, two microphones, PPG sensor, IMU, temperature sensor, NOR flash, and button.
    (d) OpenZilo worn on the finger.
    (e) Inner view of the assembled ring.
    (f) Photographs of both sides of the fabricated circuit board.
    The ring integrates speech, inertial, optical physiological, and temperature sensing within a compact wearable form factor.}
    \label{fig:ring-platform}
\end{figure}

OpenZilo integrates four sensing modalities within a compact finger-worn form factor: microphones for speech acquisition, an inertial measurement unit (IMU) for hand-motion sensing, a photoplethysmography (PPG) sensor for optical physiological measurements, and a temperature sensor for skin-temperature monitoring. These modalities provide complementary information for explicit robot commands and potential physiological-context interaction. Figure~\ref{fig:ring-platform} illustrates the mechanical assembly, circuit-board layout, and physical prototype, while Table~\ref{tab:ring_bom} summarizes the main electronic components.

\begin{table}[t]
    \centering
    \caption{Component selection for the OpenZilo prototype.}
    \label{tab:ring_bom}
    \small
    \begin{tabular}{@{}llc@{}}
        \toprule
        Component & Part number / Specification & Qty. \\
        \midrule
        MCU                & Nordic nRF54L15-CAAA  & 1 \\
        IMU                & TDK ICM-42688-P       & 1 \\
        PPG sensor         & ADI MAXM86161         & 1 \\
        Microphone         & Knowles SPH0641LU4H-1 & 2 \\
        Temperature sensor & TI TMP119             & 1 \\
        NOR flash          & 128~Mbit              & 1 \\
        \bottomrule
    \end{tabular}
\end{table}

\noindent\textbf{Processing and communication.}
OpenZilo uses the Nordic nRF54L15-CAAA microcontroller as its central processing and communication unit. The MCU coordinates sensor acquisition, local device operations, and Bluetooth Low Energy (BLE) communication with the host. A 128-Mbit NOR flash provides nonvolatile storage for recorded data. Together, the controller, storage, and wireless interface establish a shared acquisition pathway through which sensor streams can be accessed by the host SDK.

\noindent\textbf{Speech and gesture sensing.}
Two Knowles SPH0641LU4H-1 digital microphones are integrated into the ring to capture speech input. As shown in Figure~\ref{fig:ring-platform}(b), the microphones are positioned at separate locations on the circuit board. The captured audio provides input to host-side speech recognition and command mapping.

Hand motion is captured using a TDK ICM-42688-P six-axis IMU, which combines three-axis acceleration and three-axis angular velocity measurements. The IMU provides motion sequences for gesture segmentation, feature extraction, and recognition. Recognized speech and gesture events are subsequently interpreted by the application agent before being translated into robot commands.

\noindent\textbf{Physiological and temperature sensing.}
For optical physiological sensing, OpenZilo incorporates the ADI MAXM86161 PPG sensor, which integrates green, red, and infrared light-emitting diodes and a photodiode. The sensor provides an optical measurement channel for investigating physiological signals in the finger-worn form factor. A TI TMP119 temperature sensor provides an additional channel for monitoring local skin-temperature variations.These sensing channels extend OpenZilo beyond explicit speech and gesture commands toward potential physiological-context interaction.

\noindent\textbf{Mechanical integration and physical implementation.}
Figure~\ref{fig:ring-platform}(a) presents the exploded mechanical structure of OpenZilo, comprising an external button, ring housing, circuit board, curved battery, and housing cover. The curved battery and circuit-board assembly are arranged within the annular enclosure to accommodate the finger-worn form factor. The circuit board distributes components across its two sides, as illustrated in Figure~\ref{fig:ring-platform}(b). The MCU, microphones, and PPG sensor are placed on one side, while the IMU, temperature sensor, NOR flash, and button are located on the opposite side. This arrangement accommodates the four sensing modalities and supporting electronics within the available ring geometry. Figure~\ref{fig:ring-platform}(c) shows the assembled prototype worn on a finger, Figure~\ref{fig:ring-platform}(d) presents the inner ring structure, and Figure~\ref{fig:ring-platform}(e) shows the fabricated circuit board. The ring form factor introduces practical considerations including optical coupling to the finger, microphone exposure, wearing orientation, and motion artifacts, which must be considered when interpreting acquired signals.

By integrating speech, inertial, PPG, and temperature sensing into a shared hardware platform, OpenZilo provides a common acquisition interface for different robot-interaction applications.

\begin{figure}[t]
  \centering
  \includegraphics[width=0.7\linewidth]{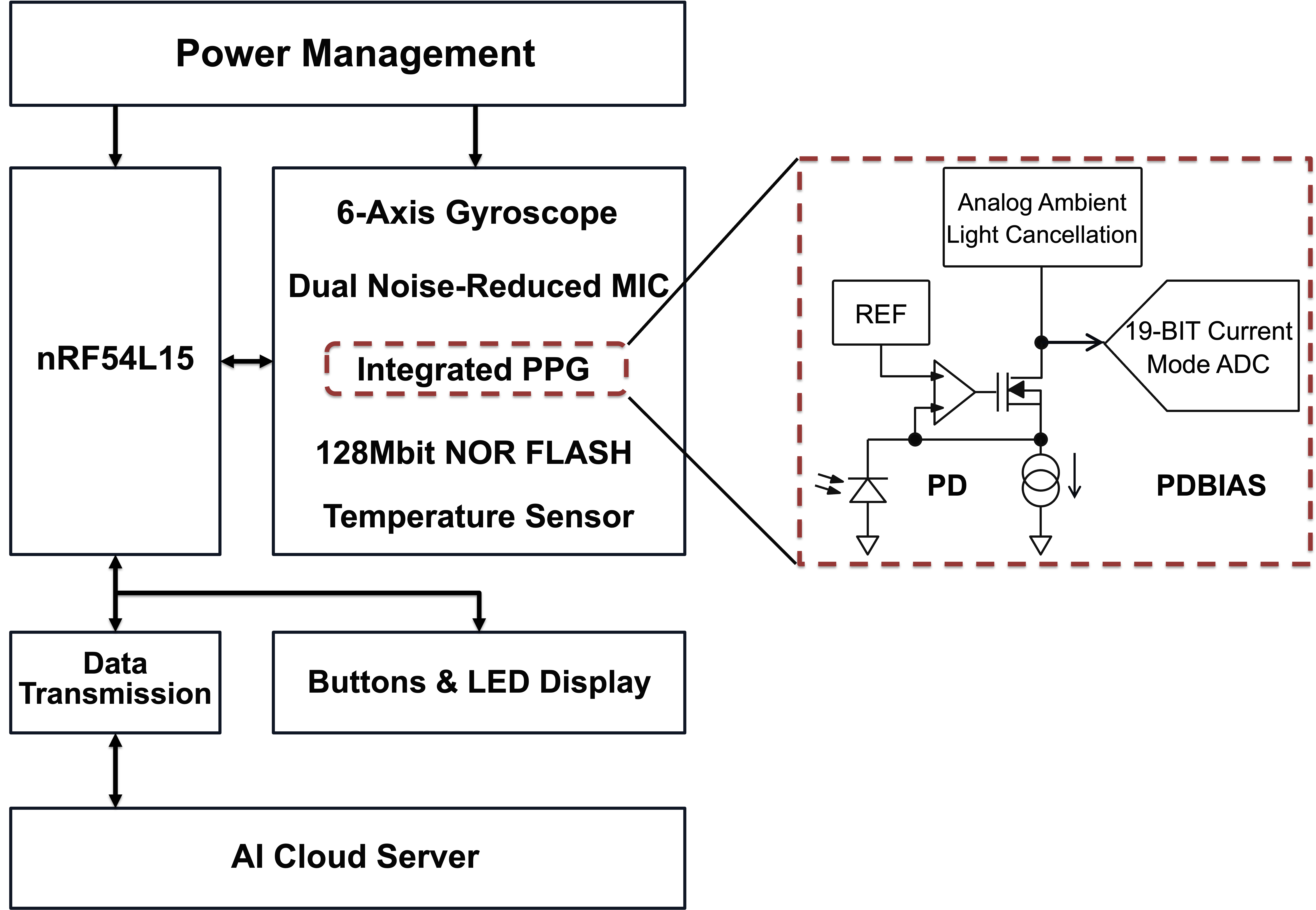}
  \caption{\textbf{Hardware architecture design schematic.} The labeled nRF54L15 controller coordinates microphone, IMU, PPG, and temperature channels with nonvolatile storage, power management, button and LED feedback, and wireless transmission. The PPG inset details photodiode bias, ambient-light cancellation, and a current-mode analog-to-digital converter (ADC). The architecture separates wearable acquisition from host processing and application logic; the downstream AI block may run at the edge or in a cloud service according to the application deployment.}
  \label{fig:hardware-scheme}
% Figure description: A block diagram connecting the nRF54L15 controller to sensors, storage, power management, local feedback and wireless host processing.
\end{figure}

\subsection{Acquisition, Power, and Communication}
The acquisition path connects ring-side buffering and storage to wireless transfer and the host SDK. The SDK separates transport from audio and IMU data paths, providing access to live streams and stored recordings. Capability checks account for differences among device and firmware versions. The rechargeable battery has a nominal capacity of 19.5 mAh; Section~\ref{sec:platform-characterization} reports current draw for the recorded operating modes. Concurrent four-sensor power, signal quality, and calibrated cross-stream alignment fall outside the present characterization.

\subsection{Open Hardware and Extensibility}
The public repository for the proposed platform provides a Python SDK, command-line tools, documentation, and device-access examples. Its software and documentation carry MPL 2.0, with a CERN-OHL-W-2.0 hardware license declaration. The SDK provides the entry point for application agents and robot adapters. Reproducibility also depends on hardware design files, assembly and firmware information, versioned models, and the robot configuration used in each study. The interaction results characterize the tested configuration; portability across sensor interfaces and robot adapters requires separate verification.

\section{SDK and Agent Integration}
\label{sec:sdk-agent}
\subsection{SDK Boundary and Perception Inputs}
The SDK described in Section~\ref{sec:platform} provides device access, live and recorded audio, audio decoding, IMU samples and calibration, and device tap and gesture events. It also exposes capability, timing, and error information. The application layer maps audio and inertial input to command events and checks them against task state before dispatch. Physiological-event notifications provide contextual inputs to the interaction policy. In RQ3, these notifications are generated manually to simulate physiological events and examine the resulting robot response.

\noindent\textbf{Speech input and command mapping.} The speech branch maps Mandarin utterances to a bounded command vocabulary. RQ1 uses forward, backward, turn left, turn right, and STOP; RQ2 uses a shared task-start instruction followed by condition-specific adjustments. A locally deployed Whisper base model~\cite{radford2023whisper} transcribes decoded microphone audio, and its output is mapped to the corresponding command event. The agent then applies task-state checks before requesting execution.

RQ1 latency starts at speech offset and excludes participant reaction and utterance duration. RQ2 response time starts at cue presentation and includes both components. Timing ends at observed motion onset for movement commands or verified stationary state for STOP. These definitions distinguish the post-input response from the complete interaction interval.

\begin{figure}[t]
\centering
\includegraphics[width=\textwidth]{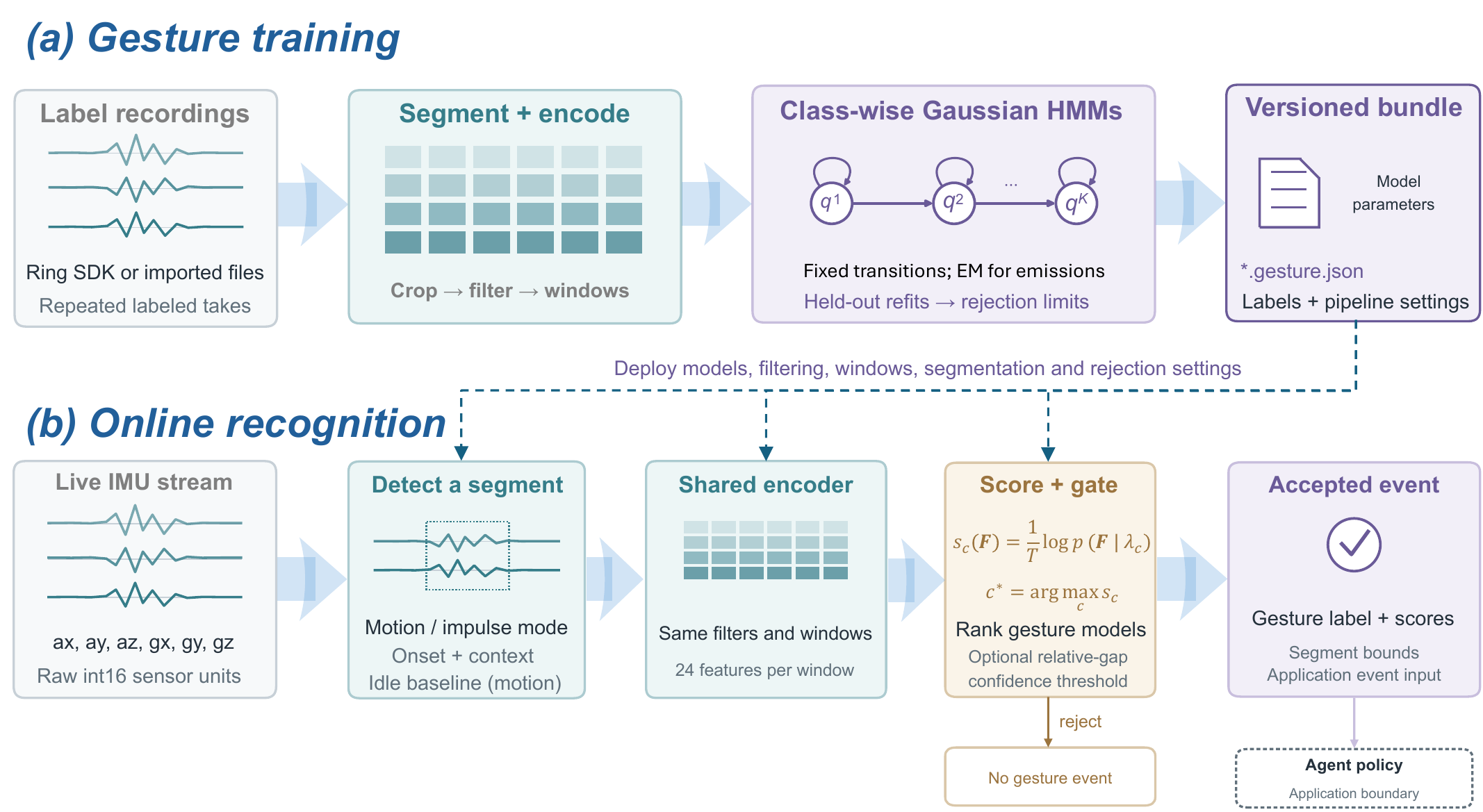}
\caption{\textbf{Host-side HMM reference pipeline for gesture training and recognition.} (a) Offline training processes labeled IMU recordings into window-level features to train gesture-specific Gaussian hidden Markov models (HMMs), which are exported together with preprocessing and recognition settings. (b) Online recognition segments live IMU signals, extracts features, and applies HMM-based scoring and rejection criteria to generate gesture events for the application agent.}
\label{fig:gesture-pipeline}
% Figure description: A two-part pipeline with offline IMU preprocessing and Gaussian HMM training, followed by online segmentation, feature extraction, model scoring and gesture-event generation.
\end{figure}

\subsection{Gesture Training and Recognition Pipeline}
\label{sec:gesture-pipeline}
The host-side HMM reference pipeline separates offline training from online event generation using six-axis IMU data. Its configuration specifies wearing orientation, calibration, preprocessing, segmentation, and feature extraction. Labeled recordings provide feature sequences for gesture-specific Gaussian HMMs. At runtime, model scoring assigns a completed segment to a gesture or rejects it. The agent determines whether an accepted event can request an action in the current task state.
Figure~\ref{fig:gesture-pipeline} shows the training and recognition path. The offline evaluation compares Gaussian HMMs with a feature-based support vector machine (SVM) across three vocabulary sizes. This comparison is separate from the five-command robot-execution profile. Gaussian HMMs were also used for gesture recognition in the command-execution study (Table~\ref{tab:rq1-gesture-execution}) and the online-adjustment study (Table~\ref{tab:rq2-summary}).

The model artifact and runtime settings jointly define recognition behavior. Calibration, feature construction, segmentation, and rejection rules can change the events delivered to the agent even when the classifier is unchanged. Reproducing a deployment therefore requires these settings alongside the trained model. The evaluation separates model-scoring time from the physical response of the robot.

\subsection{Agent State and Interaction Policy}
The agent mediates between recognized events and robot commands using event quality, freshness, task state, and physical feedback. In RQ2, the speech recognizer emits a task-start event that places the agent in an active-task state. An accepted STOP event then preempts the task, subject to recognition and state checks. Repeated or stale events must not initiate further actions. Section~\ref{sec:rq2-evaluation} specifies the task and adjustment conditions.
Physiological-event notifications provide contextual inputs to the agent, which maps an event to a task-specific response. The robot adapter translates the resulting action request into robot commands. In RQ3, the response consists of moving to a marked destination and stopping.

\subsection{Robot Adapter and Feedback}
The robot adapter translates bounded action requests into robot SDK commands and returns execution status and observed state. This feedback distinguishes software acceptance from successful physical execution. Linking acquisition records, recognized events, and robot commands supports tracing each physical outcome to its input and policy decision. The evaluation scores motion onset and stationary state as physical endpoints, as defined in Section~\ref{sec:integration-validation}.

\section{Evaluation}
\label{sec:evaluation}
The evaluation combines hardware characterization with three robot-interaction studies. RQ1 examines speech and gesture execution, together with an offline comparison of gesture models. RQ2 examines online adjustments after a shared spoken start. RQ3 examines physiological-event-triggered robot interaction through simulated event notifications and compares robot and on-site human response times.

\begin{table}[t]
  \caption{Reported current draw and idealized estimates for a nominal 19.5 mAh battery; these are not measured depletion lifetimes.}
  \label{tab:power-characterization}
  \centering
  % \small
  \begin{tabular}{lcc}
    \toprule[1.5pt]
    Operating mode & Current & $T_{\mathrm{est}}$ \\
    \midrule
    BLE disconnected & 59--61 $\mu$A & 320--331 h \\
    BLE connected & 54--55 $\mu$A & 355--361 h \\
    IMU only & 1.12 mA & 17.41 h \\
    Real-time recording & 3.88 mA & 5.03 h \\
    Non-real-time recording & 4.10 mA & 4.76 h \\
    IMU + real-time recording & 4.63 mA & 4.21 h \\
    IMU + non-real-time recording & 4.37 mA & 4.46 h \\
    \bottomrule[1.5pt]
  \end{tabular}
\end{table}

\subsection{Power and Recording Characterization}
\label{sec:platform-characterization}
Table~\ref{tab:power-characterization} reports current draw across seven operating modes. For current $I$ in mA, the capacity-based runtime estimate is $T_{\mathrm{est}}=19.5\,\mathrm{mAh}/I$, assuming constant current and full use of the nominal battery capacity. Current values in $\mu$A are converted to mA before this calculation. Bluetooth Low Energy (BLE) disconnected and connected modes yield estimates of 13.3--13.8 and 14.8--15.0 days, respectively. The table reports current ranges rather than between-run variability. The mode labels identify the recorded configurations but do not isolate contributions from storage, transmission, PPG, or skin-temperature sensing.

Two recording sessions lasted 10 h 40 min 30.2 s in IMU-only mode and 4 h 35 min 21.7 s in real-time recording mode. Their starting charge and depletion endpoints were not documented, so these durations do not measure full-charge endurance. Charging indications were sampled approximately every 5 min, using the first display of 4\% as time zero. The display reached 91\% after approximately 20 min and first showed 100\% after approximately 30 min. The charging flag first became 0 after approximately 55 min. These observations describe displayed charging milestones rather than a verified empty-to-full charging cycle.

The electrical observations characterize the listed modes and recording sessions. Simultaneous four-channel acquisition, cross-stream alignment, and valid-signal coverage require separate characterization. Capacity-based estimates and session duration should therefore be interpreted separately from measured depletion lifetime.

\subsection{Experimental Setup and Reporting}
\label{sec:integration-validation}

\begin{figure}[t]
  \centering
  \includegraphics[width=0.90\linewidth]{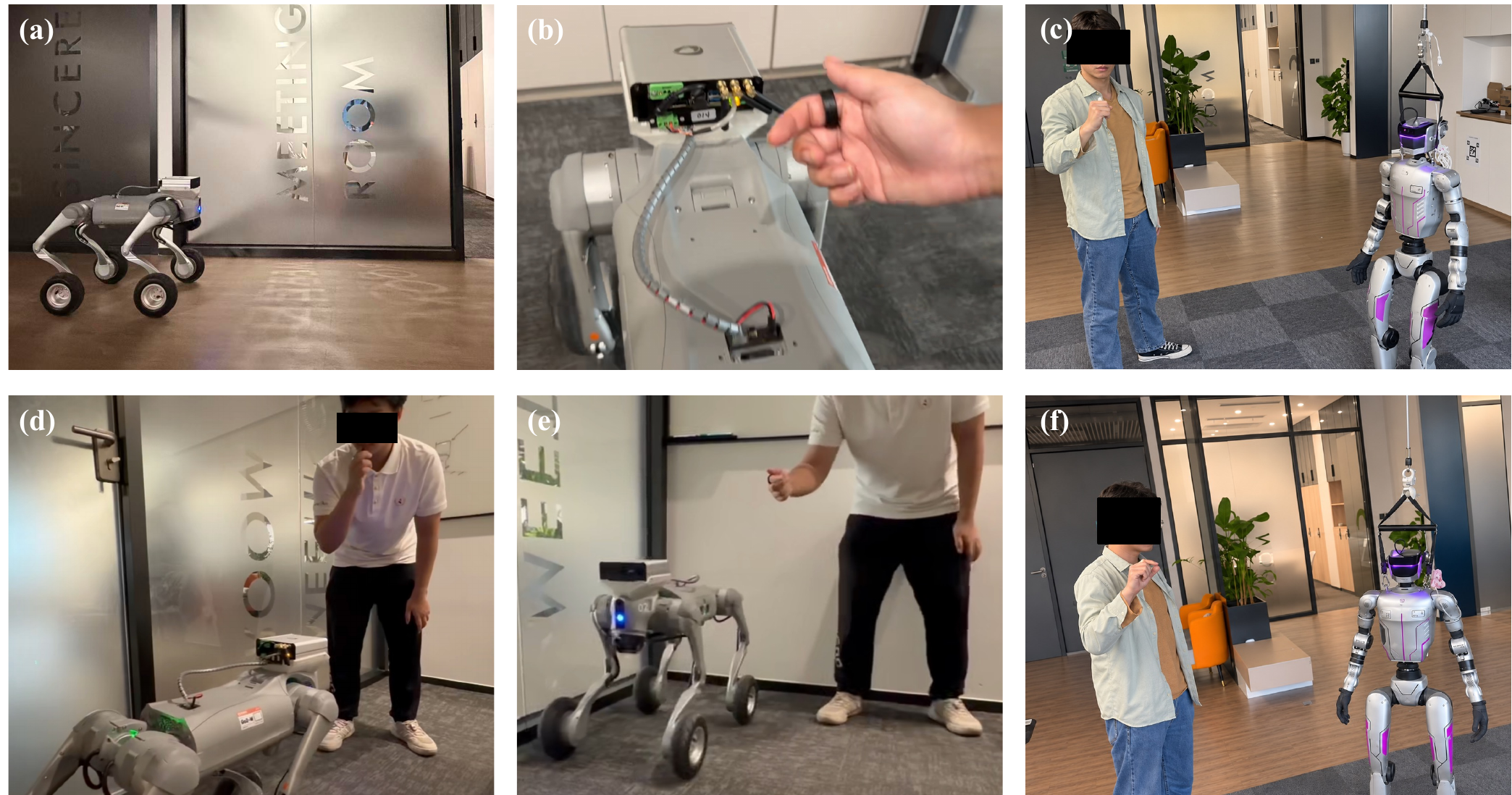}
  \caption{\textbf{Robot-control setup and interaction examples for the proposed platform.} (a) Wheeled quadruped robot. (b) Close-up of the ring-based control setup and robot-mounted hardware. (d) Speech control of the quadruped through the ring microphone. (e) IMU-based gesture control of the quadruped. (c) and (f) Extension of the proposed platform to a humanoid robot. }
\label{fig:example}
\end{figure}

Figure~\ref{fig:example} shows the wheeled quadruped setup and ring-based robot interaction examples. Panels (a) and (b) show the robot and a close-up of the control setup. Panels (d) and (e) illustrate speech and IMU-based gesture control of the quadruped, respectively. Panels (c) and (f) show an extension of the proposed platform to a humanoid robot.

\noindent\textbf{Participants, recruitment, and consent.} We recruited a convenience sample of 10 laboratory members for the command and online-adjustment studies (RQ1 and RQ2). 

We recruited a convenience sample of 10 laboratory members (8 male and 2 female), aged 20--35 years, to evaluate OpenZilo-based robot interaction. All participants wore OpenZilo on their index finger and performed speech- and gesture-based robot control tasks to assess interaction performance across different users. These studies evaluated non-clinical interaction with bounded robot routines. Physiological adaptation was disabled during the command and online-adjustment tasks. RQ3 used manually triggered notifications to simulate physiological events and compare robot and on-site human responses in a local-arrival task. All participants provided informed consent. 

\noindent\textbf{Study design and trial counts.} The experimental setup included two robotic platforms: the Unitree Go2-W, a wheeled quadruped robot, and the Unitree G1, a humanoid robot. The Go2-W served as the primary platform for quantitative interaction evaluations, while the G1 demonstrated the extensibility of OpenZilo across different robot embodiments. RQ1 and RQ2 compared input conditions within participants. Condition order was randomized independently for each participant. In RQ1, the 10 participants contributed 50 trials per command and modality across five commands, yielding 250 reported trials per modality. In RQ2, they contributed 50 task attempts per condition, each with four turn cues and one terminal STOP cue. This yielded 200 turn cues and 50 STOP cues per condition. These totals are pooled across participants; repeated trials are observations rather than independent participant samples. Some STOP observations contributed to both studies, so the reported totals do not represent entirely distinct trials. Their summaries are reported separately using study-specific timing origins.

Recognition correctness denotes a correct command label; task success additionally requires the intended physical response. Wrong actions are unintended executions, excluding failures such as no output or no action. Response times for successful commands and interventions are reported as medians, with the 25th and 75th percentiles in brackets. Local-arrival times are reported as mean $\pm$ standard deviation (SD). All comparisons are descriptive, with no population-level significance tests or participant-level confidence intervals.

Timing starts at speech offset or gesture completion in RQ1 and at cue presentation in RQ2. Movement-command latency ends at physical action onset, and STOP latency ends at verified stationary state. Failures remain in the RQ1 and RQ2 rate denominators; latency summaries include successful trials with observed response times. RQ3 uses a separate notification-receipt-to-arrival interval, defined in Section~\ref{sec:rq3-evaluation}.

\subsection{Single-Modality Command Study (RQ1)}
\label{sec:rq1-evaluation}
RQ1 evaluated forward, backward, turn left, turn right, and STOP in speech-only and gesture-only conditions. The core matched comparison covered forward, turn left, and turn right using the same robot routines. Speech commands were given in Mandarin. Forward and backward gestures involved finger extension and hand retraction toward the chest, respectively. Left and right used counterclockwise and clockwise circles in a vertical plane, viewed from the wearer's forward-looking perspective. A brief upward ring motion represented STOP. Forward and backward commands were relative to the robot's current heading.

Movement trials began with the robot stationary and requested one bounded action. Success required correct recognition and completion of the corresponding routine. STOP trials began during motion and succeeded only when the command brought the robot to a verified stationary state. For STOP, routine completion without a command-induced stop and experimenter intervention were not counted as successful responses. Only the assigned input modality dispatched commands, and physiological adaptation was disabled. Recognition and task-success rates used the same command-specific denominator, retaining recognition, communication, and execution failures.

\noindent\textbf{A. Speech-command execution.} Table~\ref{tab:rq1-speech-execution} reports 50 trials per command. Across the five commands, speech recognition was correct in 237/250 trials (94.8\%), and the requested task succeeded in 226/250 (90.4\%). Thus, 11 correctly recognized commands did not result in task success. Seven unintended physical actions were recorded. Command-specific median response times ranged from 1.32 to 1.61 s, measured from speech offset to motion onset or, for STOP, stationary state.

\begin{table}[t]
  \centering
  \caption{Speech-only command recognition and execution over 50 trials per command, pooled across 10 participants. Recognition accuracy and task success are reported as correct recognitions and successful tasks, respectively, divided by total trials (\%). Latency starts at speech offset and ends at action onset for movement commands or verified stationary state for STOP. For successful trials, latencies are reported as medians, with the 25th and 75th percentiles in brackets. Wrong actions are unintended executions.}
  \label{tab:rq1-speech-execution}
  \begin{tabular}{lcccc}
    \toprule[1.2pt]
    \multirow{2}{*}{\textbf{Speech command}}
    & \multicolumn{2}{c}{\textbf{Correct trials}}
    & \multirow{2}{*}{\textbf{Latency (s)}}
    & \multirow{2}{*}{\textbf{Wrong actions}} \\
    \cmidrule(lr){2-3}
    & \textbf{Recognition}
    & \textbf{Task success}
    & & \\
    \midrule
    Forward    & 48/50 (96.0\%) & 46/50 (92.0\%) & 1.37 [1.13, 1.68] & 1 \\
    Backward   & 47/50 (94.0\%) & 45/50 (90.0\%) & 1.54 [1.26, 1.87] & 2 \\
    Turn left  & 46/50 (92.0\%) & 43/50 (86.0\%) & 1.61 [1.32, 1.94] & 2 \\
    Turn right & 47/50 (94.0\%) & 44/50 (88.0\%) & 1.48 [1.21, 1.79] & 2 \\
    Stop       & 49/50 (98.0\%) & 48/50 (96.0\%) & 1.32 [1.09, 1.59] & 0 \\
    \bottomrule[1.2pt]
  \end{tabular}
\end{table}

\noindent\textbf{B. Gesture-command execution.} Table~\ref{tab:rq1-gesture-execution} reports the five gestures under a fixed wearing and calibration configuration, with 50 trials per command. Recognition was correct in 231/250 trials (92.4\%), and the requested task succeeded in 222/250 (88.8\%). Nine correctly recognized gestures did not result in task success, and eight unintended physical actions were recorded. Median response times ranged from 0.49 to 0.71 s, measured from gesture completion to the corresponding physical endpoint.

For the three core actions, recognition was correct in 141/150 speech trials (94.0\%) and 139/150 gesture trials (92.7\%). Both modalities achieved task success in 133/150 trials (88.7\%). Gestures had shorter command-specific median post-input latencies. These intervals excluded utterance and gesture production, so they do not measure the complete interaction response.

\begin{table}[t]
  \centering
  \caption{Gesture-only command recognition and execution over 50 trials per command, pooled across 10 participants under a fixed wearing and calibration configuration. Recognition accuracy and task success are reported as correct recognitions and successful tasks, respectively, divided by total trials (\%). Latency starts at gesture completion and ends at action onset for movement commands or verified stationary state for STOP. For successful trials, latencies are reported as medians, with the 25th and 75th percentiles in brackets. Wrong actions are unintended executions.}
  \label{tab:rq1-gesture-execution}
  \begin{tabular}{lcccc}
    \toprule[1.2pt]
    \multirow{2}{*}{\textbf{Gesture command}}
    & \multicolumn{2}{c}{\textbf{Correct trials}}
    & \multirow{2}{*}{\textbf{Latency (s)}}
    & \multirow{2}{*}{\textbf{Wrong actions}} \\
    \cmidrule(lr){2-3}
    & \textbf{Recognition}
    & \textbf{Task success}
    & & \\
    \midrule
    Forward    & 46/50 (92.0\%) & 44/50 (88.0\%) & 0.56 [0.41, 0.73] & 1 \\
    Backward   & 45/50 (90.0\%) & 43/50 (86.0\%) & 0.62 [0.46, 0.82] & 2 \\
    Turn left  & 46/50 (92.0\%) & 44/50 (88.0\%) & 0.71 [0.54, 0.92] & 2 \\
    Turn right & 47/50 (94.0\%) & 45/50 (90.0\%) & 0.67 [0.50, 0.87] & 2 \\
    Stop       & 47/50 (94.0\%) & 46/50 (92.0\%) & 0.49 [0.36, 0.65] & 1 \\
    \bottomrule[1.2pt]
  \end{tabular}
\end{table}

\noindent\textbf{C. Gesture-model comparison across vocabulary sizes.} The offline evaluation compared a feature-based SVM with Gaussian HMMs at vocabulary sizes of 4, 8, and 12 classes. An 80:20 training--test split provided 50 held-out trials per class, giving 200, 400, and 600 test trials, respectively. Model scoring ran on a workstation equipped with an NVIDIA RTX 5090. These vocabularies tested recognition scaling separately from the five-command robot-execution profile.

Each complete gesture segment contributed one prediction. Accuracy counted rejection and no output as errors; Macro-F1 averaged F1 over the gesture classes. Rejection and no output contributed false negatives for the true class. Model-scoring time was the median time per complete gesture segment, excluding acquisition, feature extraction, communication, and robot motion. This interval is distinct from the physical response time measured in the execution study.

Table~\ref{tab:gesture-training-capacity} shows higher observed accuracy and Macro-F1 for Gaussian HMMs at every vocabulary size. The accuracy difference relative to SVM ranged from 3.5 to 4.0 percentage points. From 4 to 12 classes, HMM accuracy declined from 95.0\% to 88.2\%, and median scoring time increased from 3.6 to 10.2 ms. SVM accuracy declined from 91.5\% to 84.2\%, while scoring time increased from 1.3 to 2.1 ms. These results quantify the accuracy--computation tradeoff for the tested vocabularies.

\begin{table}[t]
  \centering
  \caption{Gesture recognition across vocabulary sizes with an 80:20 training--test split and 50 held-out test trials per class. Vocabulary size is denoted by $K$; $k/N$ reports correct predictions over all test trials, with rejection and no output counted as errors. Macro-F1 is averaged over gesture classes. Scoring time is the median per complete segment on a workstation equipped with an NVIDIA RTX 5090.}
  \label{tab:gesture-training-capacity}
  \begin{tabular}{clccc}
    \toprule[1.2pt]
    \multirow{2}{*}{$K$}
    & \multirow{2}{*}{\textbf{Algorithm}}
    & \multicolumn{2}{c}{\textbf{Recognition}}
    & \multirow{2}{*}{\textbf{Scoring time (ms)}} \\
    \cmidrule(lr){3-4}
    & & \textbf{$k/N$ (\%)}
    & \textbf{Macro-F1}
    & \\
    \midrule
    \multirow{2}{*}{4}  & Feature-based SVM
       & 183/200 (91.5\%) & 0.921 & 1.3 \\
       & Gaussian HMM
       & 190/200 (95.0\%) & 0.952 & 3.6 \\
    % \midrule
    \multirow{2}{*}{8}   & Feature-based SVM
       & 351/400 (87.8\%) & 0.887 & 1.6 \\
       & Gaussian HMM
       & 366/400 (91.5\%) & 0.923 & 6.7 \\
    % \midrule
    \multirow{2}{*}{12}  & Feature-based SVM
       & 505/600 (84.2\%) & 0.853 & 2.1 \\
       & Gaussian HMM
       & 529/600 (88.2\%) & 0.891 & 10.2 \\
    \bottomrule[1.2pt]
  \end{tabular}
\end{table}

\subsection{Spoken Task and Online Adjustment Study (RQ2)}
\label{sec:rq2-evaluation}
Both conditions began with the same Mandarin instruction to start a predefined low-speed forward task. Condition V used speech for online left/right turns and STOP; condition G used the ring gestures described in RQ1. The robot routines were shared across conditions. A turn interrupted translation, rotated in place, and resumed forward motion along the new heading. STOP ended the task without automatic resumption. Physiological adaptation was disabled.

Each condition comprised 50 task attempts, with four turn cues and one terminal STOP per trial. The denominators were 200 turn cues and 50 STOP cues per condition. A successful turn required rotation in the cued direction followed by task resumption; a successful STOP required a verified stationary state. Whole-task success required a successful start, all required adjustments, and terminal STOP. Response times ran from cue presentation to rotation onset or stationary state, including participant reaction, input production, processing, communication, and robot response.

\begin{table}[t]
  \centering
  \caption{Online-adjustment performance following the same spoken task start, pooled across 10 participants. V uses speech and G uses gestures for online adjustments. Each condition contains 50 task attempts, with $N_T=200$ turn cues and $N_S=50$ terminal STOP cues. Success is reported as successful events/eligible events (\%) or successful tasks/task attempts (\%). Turn latency runs from cue presentation to rotation onset; STOP latency runs from cue presentation to verified stationary state. For successful events, latencies are reported as medians, with the 25th and 75th percentiles in brackets.}
  \label{tab:rq2-summary}
  \begin{tabular}{@{}lccccc@{}}
    \toprule[1.2pt]
    \multirow{2}{*}{\textbf{Condition}}
    & \multicolumn{2}{c}{\textbf{Turn}}
    & \multicolumn{2}{c}{\textbf{STOP}}
    & \multirow{2}{*}{\textbf{Whole-task success}} \\
    \cmidrule(lr){2-3}
    \cmidrule(lr){4-5}
    & \textbf{$k/N_T$ (\%)}
    & \textbf{Latency (s)}
    & \textbf{$k/N_S$ (\%)}
    & \textbf{Latency (s)}
    & \\
    \midrule
    V & 181/200 (90.5\%) & 2.32 [1.98, 2.75]
      & 47/50 (94.0\%) & 2.18 [1.85, 2.61]
      & 32/50 (64.0\%) \\
    G & 185/200 (92.5\%) & 1.74 [1.43, 2.12]
      & 48/50 (96.0\%) & 1.24 [1.01, 1.52]
      & 35/50 (70.0\%) \\
    \bottomrule[1.2pt]
  \end{tabular}
\end{table}

Table~\ref{tab:rq2-summary} reports turn success of 90.5\% for V and 92.5\% for G, with STOP success of 94.0\% and 96.0\%, respectively. Median turn response time was 2.32 s for V and 1.74 s for G; median STOP response time was 2.18 s and 1.24 s. The gesture condition had shorter observed medians by 0.58 s for turning and 0.94 s for stopping. Whole-task success was 32/50 (64.0\%) for V and 35/50 (70.0\%) for G. These differences describe the tested configuration.

Some STOP observations also contributed to the RQ1 execution profiles. The studies use different timing origins, and their STOP summaries are not independent replications. Each result is reported using its study-specific definition, without pooling overlapping observations.

\subsection{Physiological-Event-Triggered Robot Interaction (RQ3)}
\label{sec:rq3-evaluation}
RQ3 examines how physiological-event notifications initiate robot actions and compares robot and on-site human response times under simulated scenarios. The evaluated response consists of moving to a marked destination and stopping. Implications for physiological-event inference and user control are discussed in Section~\ref{sec:event-discussion}.

\textbf{A. Physiological-event simulation and response task.} Notifications were manually triggered to simulate physiological events. A quadruped robot and an on-site person responded under matched route and starting-distance conditions. Following notification receipt, each responder moved to a marked destination and stopped. Arrival was defined as reaching the destination and coming to a stop. This procedure evaluated the response to an event notification without requiring an automatic physiological-event detector.

Response time was measured from verified notification receipt to arrival, excluding the interval between manual triggering and notification receipt. Table~\ref{tab:notification-arrival-local} summarizes 28 valid timing records per responder under the simulated physiological-event scenarios. Mean response times were $11.71 \pm 2.94$ s for the robot and $9.20 \pm 2.56$ s for the on-site person. The robot's mean interval was 2.51 s longer under the tested conditions. Note that these descriptive summaries characterize the recorded arrivals; they do not estimate arrival success across all attempted notifications.

\begin{table}[t]
  \centering
  \caption{Robot and on-site human response times following manually triggered physiological-event notifications in simulated scenarios. Times run from verified notification receipt to arrival at the marked destination and stopping. Values are mean $\pm$ SD over $n=28$ valid timing records per responder under matched route and starting-distance conditions.}
  \label{tab:notification-arrival-local}
  \begin{tabular}{lcc}
    \toprule[1.2pt]
    \textbf{Responder}
    & \textbf{Valid timing records ($n$)}
    & \textbf{Receipt$\to$arrival (s)} \\
    \midrule
    Quadruped robot
    & 28
    & $11.71 \pm 2.94$ \\

    On-site person
    & 28
    & $9.20 \pm 2.56$ \\
    \bottomrule[1.2pt]
  \end{tabular}
\end{table}

\textbf{B. External EMS response context.} Table~\ref{tab:ems-response-context} provides two non-fall-specific US EMS summaries as external context. Mell et al. reported response times from 911 call receipt to first-unit arrival for 1,753,168 encounters across 485 agencies in 2015~\cite{mell2017response}. Hayes et al. summarized 32,247,001 activations in 2023, measuring dispatch to arrival at the patient's side~\cite{hayes2026response}. Although differences in timing definitions, endpoints, populations, and summary statistics preclude direct comparisons with local robot arrival, these findings highlight the potential of robot-mediated interaction to enable rapid, localized responses to physiological events.

\begin{table}[t]
  \centering
  \caption{Published US EMS response-time statistics. Reported intervals, populations, and summary measures differ across sources and are not directly comparable with each other or with the local-arrival intervals in Table~\ref{tab:notification-arrival-local}.}
  \label{tab:ems-response-context}
  \begin{tabularx}{\columnwidth}{@{}p{0.17\columnwidth}X p{0.20\columnwidth}p{0.18\columnwidth}@{}}
    \toprule[1.2pt]
    \textbf{Scope}
    & \textbf{Response interval}
    & \textbf{Time (min)}
    & \textbf{Source} \\
    \midrule
    US, 2015
    & 911 receipt $\to$ first EMS unit on scene
    & $7.9 \pm 5.7$ (mean $\pm$ SD)
    & Mell et al.~\cite{mell2017response} \\
    US, 2023
    & Dispatch $\to$ patient's side
    & 7.1 (median)
    & Hayes et al.~\cite{hayes2026response} \\
    \bottomrule[1.2pt]
  \end{tabularx}
\end{table}

\section{Discussion and Limitations}
\label{sec:discussion}
\subsection{Tracing Sensing to Robot Action}
The proposed platform connects wearable recognition to physical execution through explicit software boundaries. RQ1 demonstrates the importance of evaluating both: 11 correctly recognized speech commands and nine correctly recognized gestures did not produce successful tasks. Recognition accuracy alone would omit these failures. Separating recognition, policy, dispatch, and feedback provides a structure for locating failures, although the aggregate results do not identify their individual causes.

This separation supports research beyond classifier selection. Researchers can vary rejection rules, task-state constraints, or robot adapters and assess both accepted events and their physical consequences. Linked event and action records, together with consistent success criteria, are necessary for such comparisons. The proposed platform thus provides a common experimental structure for studying how perception and control jointly affect interaction outcomes.

\subsection{Comparing Input across Task Stages}
The results motivate evaluating modality choice at the task stage where it is used. Both modalities achieved 133/150 task successes on RQ1's three core actions. In RQ2, gestures had shorter observed cue-to-turn and cue-to-stop times and three additional successful tasks. This pattern supports further investigation of spoken task initiation combined with gesture-based adjustment. The 10-participant sample and descriptive analysis limit generalization beyond the tested configuration.

The timing distinction is central to this interpretation. RQ1 starts after the input ends, RQ2 includes reaction and input production, and the offline benchmark measures model scoring alone. These intervals address different questions about recognizer computation, post-input response, and the complete interaction. A reduction in one interval therefore need not imply an equivalent reduction in another.

\subsection{Physiological-Event-Triggered Interaction and User Control}
\label{sec:event-discussion}
RQ3 connects physiological-event notifications to a concrete physical response through the platform's event-to-action interface. The simulated scenarios characterize the response stage after notification receipt and compare robot and on-site human arrival times. Extending this interaction to sensor-generated notifications requires validated event inference and measurement of the complete sensing-to-action interval.

Sensor-driven pause-and-resume interaction provides a further extension of the event-to-action pathway. PPG and temperature could inform a bounded PAUSE only after signal-quality checks and independent validation of the event definition. PAUSE would retain the task, whereas STOP ends it; resumption would require both valid event clearance and explicit user confirmation. Future evaluation should measure incorrect interruptions, missed events, pause duration, and recovery burden alongside task completion. The current command and arrival results do not validate these physiological inferences or adaptation policies.

\subsection{Reproducibility and Extension}
The offline comparison shows why model selection should account for both recognition and interaction requirements. Gaussian HMMs had higher observed accuracy and Macro-F1 than SVM across the tested vocabularies, with longer scoring times. Both models lost accuracy as vocabulary size increased. These observations motivate evaluating each proposed vocabulary against both recognition accuracy and response-time requirements. An operational vocabulary limit would also require acceptance criteria and continuous-activity measurements of unintended activation.

Reproducible extension requires sensing settings, trained models, event definitions, and robot mappings to be versioned together. Researchers can then compare component changes while preserving the rest of the configuration. A replacement robot adapter must preserve action semantics and physical outcome checks. Cross-laboratory reuse and reductions in development effort remain separate evaluation questions.

\subsection{Limitations and Validation Scope}
The command studies involved 10 participants performing repeated, bounded tasks. Pooled summaries describe these observations but do not quantify participant-level uncertainty or establish performance during everyday use. The 80:20 gesture split alone does not establish generalization to unseen participants or sessions. The offline benchmark also lacks a five-class configuration matched to the execution profile, so its results cannot substitute for deployment-specific evaluation.

Response-time summaries are conditional on successful execution and must be interpreted alongside failure rates. Wrong-action counts do not measure unintended activation during unprompted activity, for which no exposure-based results are reported. Partial reuse of STOP trials also limits independence between RQ1 and RQ2. RQ3 evaluates responses to simulated physiological-event notifications; automatic physiological-event detection and sensor-driven pause-and-resume policies remain to be evaluated. These limits constrain claims about continuous reliability and health-related assistance.

Hardware characterization covers the recorded operating modes rather than verified simultaneous PPG, temperature, audio, and IMU acquisition. Capacity-based runtime is an estimate, and the recording sessions lack documented full-charge depletion endpoints. Concurrent sensing quality, temporal alignment, and battery endurance require further measurement before extending the findings to continuous multisensor use.

\subsection{Future Work}
Future work should assess whether the observed interaction patterns persist across participants, sessions, and everyday settings. Continuous-use studies should measure unintended activations per hour, recovery effort, and battery endurance under concurrent sensing. Synchronized sensor, event, and robot-state records would help locate failures across the interaction chain. Future work should extend the simulated-event interaction to notifications derived from validated physiological sensing. Pause-and-resume policies should incorporate user confirmation before task resumption. Implementing the same action semantics on additional robots would test the portability of both the architecture and the interaction findings.

\section{Conclusion}
\label{sec:conclusion}
The proposed platform combines a four-sensor smart ring with interfaces separating acquisition, event interpretation, interaction policy, and robot feedback. The evaluation links hardware operation and gesture recognition to physical command execution, online adjustment, and physiological-event-triggered robot interaction under simulated scenarios. Recognition accuracy exceeded task success for both speech and gestures, while gesture-based online adjustments had shorter observed response times in the 10-participant study. The offline comparison showed higher accuracy and longer scoring times for HMMs, with declining accuracy as the tested vocabularies expanded. These findings motivate evaluating wearable robot interfaces through both recognition and physical outcomes, using timing definitions appropriate to each task stage. Broader participant coverage, continuous-use testing, and validated physiological events remain necessary extensions.

\section*{LLM Usage Clarification}
We used ChatGPT to edit the manuscript for grammar, clarity, and readability. We also used it to generate the illustrative character images in Figure~\ref{fig:introduction}.

\bibliographystyle{unsrtnat}
\bibliography{sample-base}

\end{document}